\documentclass[twoside,11pt]{article} 

\usepackage{jmlr2e} 
\usepackage{balance}  
\usepackage{url}      

\usepackage{times}
\usepackage{helvet}
\usepackage{courier}
\usepackage{graphicx}
\usepackage{subcaption}

\usepackage{amsfonts}
\usepackage{amsmath}
\usepackage{algorithmicx}
\usepackage{algpseudocode}
\usepackage{algorithm}

\makeatletter
\def\url@leostyle{%
  \@ifundefined{selectfont}{\def\UrlFont{\sf}}{\def\UrlFont{\small\bf\ttfamily}}}
\makeatother
\firstpageno{1}

\begin{document}

\title{Using PCA to Efficiently Represent State Spaces}

\author{\name William Curran \email curranw@onid.oregonstate.edu\\ 
\addr Oregon State University\\ 
Corvallis, Oregon
\AND
\name Tim Brys \email timbrys@vub.ac.be \\
\addr Vrije Universiteit Brussel \\
\addr Brussels, Belgium
\AND
\name Matthew E. Taylor \email taylorm@eecs.wsu.edu \\
\addr Washington State University \\
\addr Pullman, Washington
\AND
\name William D. Smart \email bill.smart@oregonstate.edu\\
\addr Oregon State University\\ 
Corvallis, Oregon
}


\maketitle

\begin{abstract}

Reinforcement learning algorithms need to deal with the exponential growth of states and actions when exploring optimal control in high-dimensional spaces. This is known as the curse of dimensionality. By projecting the agent's state onto a low-dimensional manifold, we can represent the state space in a smaller and more efficient representation. By using this representation during learning, the agent can converge to a good policy much faster. We test this approach in the Mario Benchmarking Domain. When using dimensionality reduction in Mario, learning converges much faster to a good policy. But, there is a critical convergence-performance trade-off. By projecting onto a low-dimensional manifold, we are ignoring important data. In this paper, we explore this trade-off of convergence and performance. We find that learning in as few as 4 dimensions (instead of 9), we can improve performance past learning in the full dimensional space at a faster convergence rate.
\end{abstract}

\section{Introduction}

Learning in high dimensional spaces is necessary and difficult in robotic applications. State and action spaces in robotics become large, continuous, and scale exponentially with the number of joints. This leads to the \textit{curse of dimensionality}.  

To address this issue, researchers have developed transfer learning \citep{5288526} and learning from demonstration \citep{Argall:2009:SRL:1523530.1524008} approaches. Transfer learning reduces computational complexity by learning in a simple domain, and transferring that knowledge to a more complex domain. In transfer learning there are three key research questions: What to transfer, how to transfer and when to transfer. Each of these questions are difficult to answer and all are completely domain dependent. When the source domain and the target domain are loosely related, straight forward transfer learning does not work, and can lead to worse performance \citep{5288526}. Transfer learning also requires a computable mapping from the source domain to the target domain, which isn't always possible.

Learning from demonstration (LfD) methods speed up convergence time by bootstrapping learning with  demonstrations \citep{Argall:2009:SRL:1523530.1524008}. LfD learns a policy using examples or demonstrations provided by a human or a robotic teacher. This method extracts state-action pairs from these demonstrations to bootstrap learning.  However, these demonstrations must be consistent and accurately represent solving the task. These methods also solve a specific complex task, rather than solve for general control \citep{Argall:2009:SRL:1523530.1524008}.

In this work, we focus on the core problem of dealing with large state spaces. We approach this issue by projecting the full state space onto a low-dimensional manifold. We use Principal Component Analysis to find this transform, and use it during each learning iteration. In this way, we perform learning in only the low-dimensional space. This leads to a critical trade-off. By projecting onto a low-dimensional manifold, we are throwing out low variance, yet potentially important data. However, learning can converge to a good, yet suboptimal, policy much faster. In this paper, we explore this trade-off of convergence and performance.

We organize the rest of this paper as follows. Section 2 describes related work in dimensionality reduction in machine learning and reinforcement learning. In Section 3 we describe our work with PCA, dimensionality reduction and reinforcement learning. Section 4 introduces the Mario Benchmarking Domain and our learning approach. Experimental results are in Section 5, followed by conclusion and related work in Sections 6 and 7.

\section{Related Work}
To motivate our approach, we introduce previous work performed in the field of dimensionality reduction.

\subsection{Dimensionality Reduction}
Previous work in dimensionality reduction focuses on reducing the space for classification or function approximation. PCA is effective in many machine learning and data mining applications to extract features from large data sets \citep{Pechenizkiy:features, 139758}. Rather than using PCA for feature extraction in large data sets, we use PCA to reduce the dimensionality of the state space during learning.

\citet{Swinehart05dimensionalreduction} used a similar approach for function approximation with neural networks. They find that they can reduce convergence time for reinforcement-based, random walk learning can by reducing the dimension of the parameter space. \citet{Liu11compressivereinforcement} also use dimensionality reduction to compute policies in a low dimensional subspace. Liu computes the low dimensional subspace from a high-dimensional space through random projections. They also reduce convergence time in continuous state spaces.

\section{PCA+RL in High-Dimensional Spaces}
\label{High-Dimensional}

to be viable in many scenarios robots need to perform complex manipulation tasks. These complex manipulations need high degree-of-freedom arms and manipulators. For example, the PR2 robot has two 7 DoF arms. When learning position, velocity and acceleration control, this leads to a 21 dimensional state space per arm. Learning in these high-dimensional spaces is computationally intractable without optimization techniques.

To learn in high-dimensional state spaces, our algorithm first computes a transform between the high-dimensional state space and a lower-dimensional space. To perform this computation, we need trajectories across a representative set of the agent's state space. We can then use any dimensionality reduction technique to learn the transform. In this work, we use Principal Component Analysis (PCA) \citep{Shlens05atutorial}. 

PCA identifies patterns in data and reduces the dimensions of the dataset with minimal loss of information. It does this by computing a transform to convert correlated data to linearly uncorrelated data. This transformation ensures that the first principal component has the largest possible variance. Each additional component has the largest possible variance uncorrelated with all previous components. Essentially, PCA represents as much as the demonstrated state space as possible in a lower dimension. This transform is given by:
\begin{equation}
T = XW
\end{equation}
where $X$ is the demonstrated data, $W$ is a $p$ by $p$ matrix whose columns are eigenvectors of $X^TX$ and $p$ is the number of principle components (in this case, the number of dimensions). 

To transform to any arbitrary dimension, $k$, we can choose $k$ eigenvectors from $W$ with the largest eigenvalues to form a $p$ by $k$ dimensional matrix $W_k$:
\begin{equation}
T_k = XW_k
\end{equation}
 
We then use reinforcement learning to learn trajectories in the new manifold. All learning is in the lower-dimensional space. For each learning iteration, we project state $x$ down to the lower-dimensional space $k$:
\begin{equation}
x_k = W^T_kx
\end{equation}

\begin{figure}[h!]
  \centering
      \includegraphics[width=0.75\textwidth]{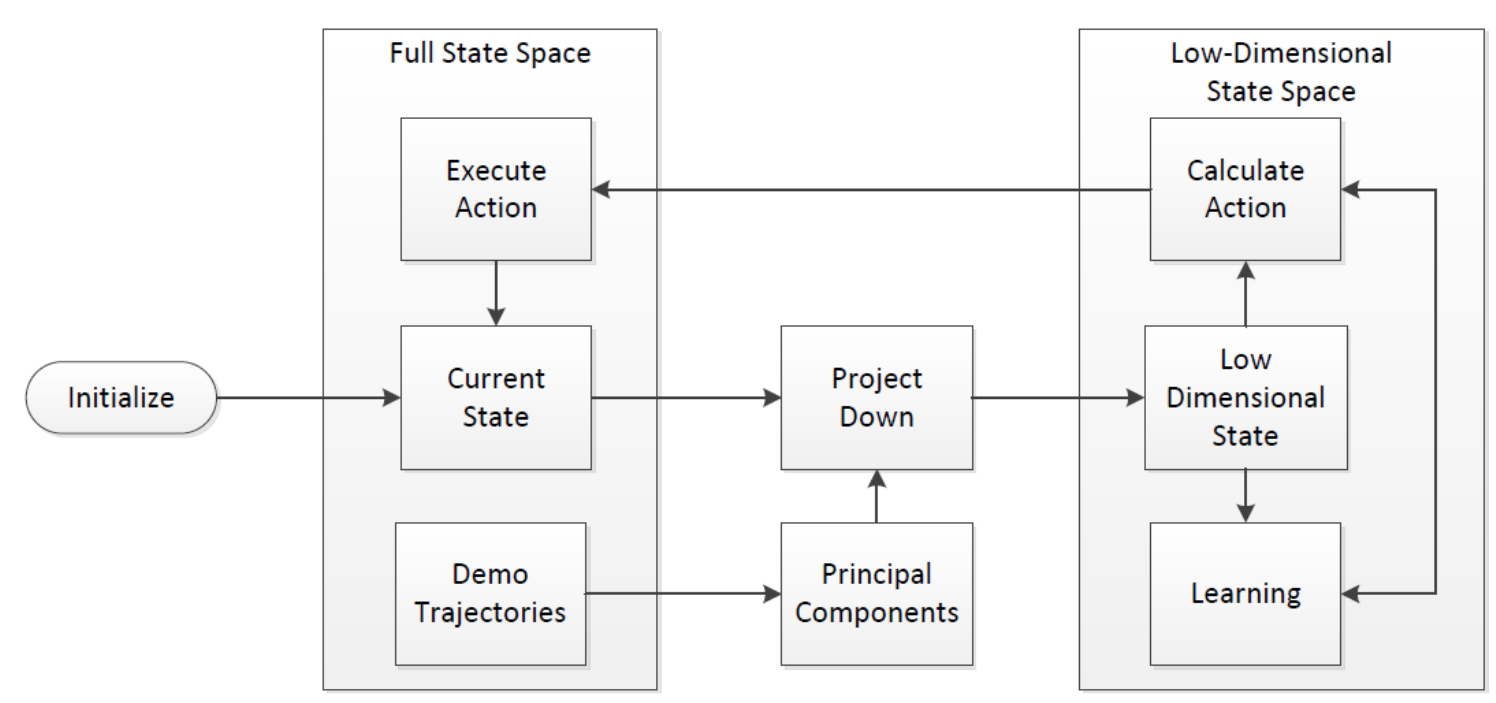}
  \caption{PCA+RL Flowchart}
  \label{fig:flowchart}
\end{figure}

We can now compute the action using the chosen learning algorithm, and execute that action in the simulation. The simulation calculates the new state given the executed action, and we can project that state down to the same lower-dimensional space.  We can then perform a learning update (Figure \ref{fig:flowchart}). 

By learning in a smaller space, reinforcement learning algorithms will converge must faster. However, PCA cannot represent all the variance in all the demonstrations. Therefore, given infinite time, the converged learning performance will always be worse than learning in the full space. This leads to a critical trade-off. By projecting onto a low-dimensional manifold, we are throwing out low variance, yet possibly important data. Yet, learning can still converge to a good, yet suboptimal, policy much faster.

\section{Mario Benchmark Problem}

\begin{figure}[h!]
  \centering
      \includegraphics[width=0.5\textwidth]{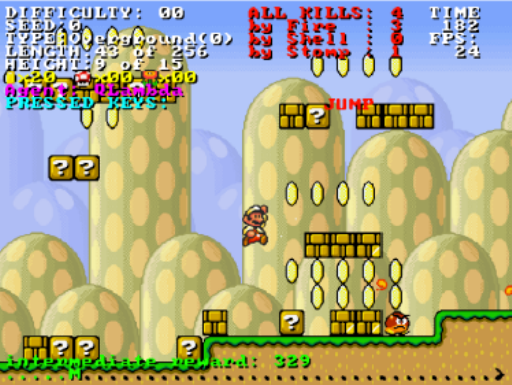}
  \caption{A screenshot of Mario.}
  \label{fig:mario}
\end{figure}

The Mario benchmark problem \citep{karakovskiy2012mario} is based on Infinite Mario Bros, which is a public reimplementation of the original $80$'s game Super Mario Bros$^\text{\textregistered}$. In this task, Mario needs to collect as many points as possible, this is done by killing an enemy ($10$), devouring a mushroom ($58$) or a fireflower ($64$), grabbing a coin ($16$), finding a hidden block ($24$), finishing the level ($1024$), getting hurt by a creature ($-42$) or dying ($-512$). The actions available to Mario correspond to the buttons on the NES controller, which are (left, right, no direction), (jump, don't jump), and (run/fire, don't run/fire). Mario can take one action from each of these groups simultaneously, resulting in $12$ distinct combined or `super' actions. The state space in Mario is quite complex, as Mario observes the exact locations of all enemies on the screen and their type. He also observes all information about himself, such as what mode he is in (small, big, fire). Lastly, he has a gridlike receptive field in which each cell indicates what type of object is in it (such as a brick, a coin, a mushroom, a goomba (enemy), etc.). A screenshot is shown in Figure~\ref{fig:mario}.
 
Our reinforcement learning agent for Mario is inspired by Liao's and Brys' previous work \citep{brys2014multi,liao2012cs229}. We use a $Q(\lambda)$-learner with a tabular state representation and $\alpha$=0.01, $\lambda$=0.5, $\gamma$=0.9 and $\epsilon$=0.05. The part of the state space the agent considers consists of these variables:

\begin{itemize}
\item is Mario able to jump ($0-1$)
\item is Mario on the ground ($0-1$)
\item is Mario able to shoot fireballs ($0-1$)
\item Mario's current direction, 8 directions and standing still ($0-8$)
\item enemies closeby (within one gridcell) in 8 directions ($2^8 \rightarrow 0-255$)
\item enemies at midrange (within one to three gridcells) in 8 directions ($0-255$)
\item whether there is an obstacle in four vertical gridcells in front of Mario ($2^4 \rightarrow 0-16$)
\item closest enemy position within 21x21 grid surrounding Mario + 1 for absent enemy ($21^2+1 \rightarrow 0-441$)
\end{itemize}

This makes for $3.24 \times 10^{10}$ possible states, and $4 \times 10^{11}$ $Q$-values ($12$ actions in each state). The size of the state space is not a problem computationally, because this state-space is sparsely visited, with the majority of the state-visits to a small set of states \citep{liao2012cs229}.

Previous work has investigated using demonstrations in the Mario domain to speed up reinforcement learning, albeit in different ways: by shaping the reward using the demonstration \citep{Brys2015IJCAI}, or learning a reward function using inverse reinforcement learning \citep{lee2014learning}.

In the experiments, we run every learning episode on a procedurally generated level based on a random seed $\in [0,10^6]$, with difficulty $0$. Also, we randomly select the mode Mario starts in (small, large, fire) for each episode. Making an agent learn to play Mario this way helps avoid overfitting on a specific level, and makes for a more generally applicable Mario agent. Our results are always averaged over $100$ different trials.

\section{Results}

As a preliminary analysis, we calculated all the principle components of the Mario domain to see which dimensions PCA weighed the highest during learning. The \textit{jump}, \textit{ground} and \textit{current direction} features are heavily represented in the first few principal components. This is intuitive, as this state changes frequently throughout a game of Mario. These features are also fundamental skills required to play a game of Mario. 

PCA also associated features related to enemies within close proximity to Mario entirely in the last few principal components. These features are only important in very specific scenarios where Mario needs to quickly react to many nearby enemies. If only one enemy is nearby, it is also represented in the \textit{closest enemy X} and \textit{closest enemy y} features.

This analysis legitimizes our approach in the Mario Benchmarking Domain. It demonstrates that we should initially learn using the fundamental skills required to play Mario. We will show that we can learn these skills quickly. The skills represented in the higher principal components are better for strict optimization in specific scenarios. Analyzing the principal components of the demonstrations is a sanity check as well as validation of the richness of the demonstration.

\begin{figure}[h!]
  \centering
      \includegraphics[width=.75\textwidth]{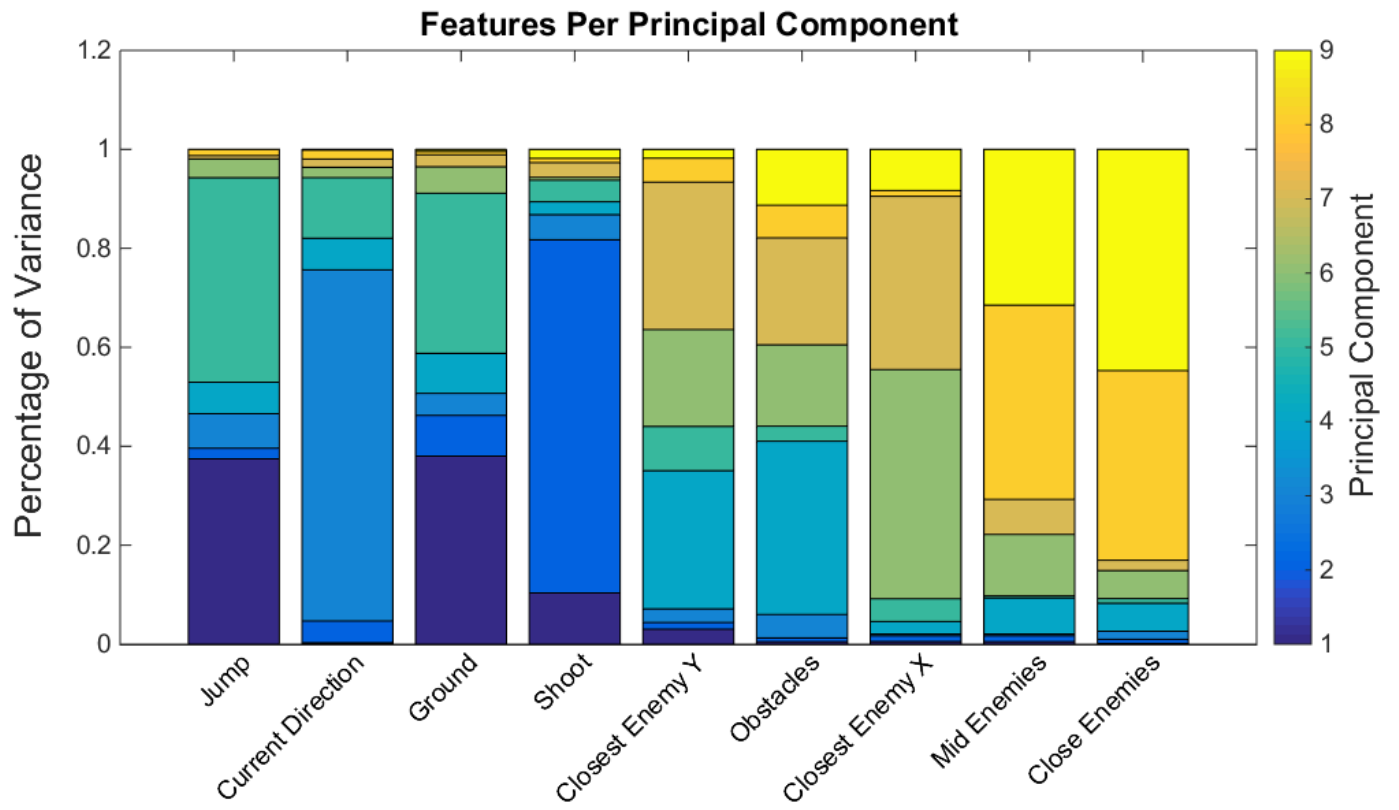}
  \caption{The emphasis of each feature relative to the principal component.}
  \label{fig:pca}
\end{figure}

When using our approach in the Mario domain, results were as expected. By projecting the state down to a low-dimensional manifold (less than 4), the learning algorithm converged quickly to bad policies. However, when using a manifold of 4 dimensions or greater, we converged quickly to a much better policy. These are promising results although these well-performing dimensions may still converge to a suboptimal policy after more than 5000 episodes.

Since learning was poor in the first two manifolds, it shows us that the \textit{jump} and \textit{ground} features are not informative enough alone to learn an effective Mario policy. Yet, when projecting onto a 3 or 4 dimensional manifold, we see large increase in policy performance. In these manifolds, the features \textit{jump}, \textit{ground}, \textit{current direction}, \textit{shoot}, \textit{closest enemy Y}, and \textit{obstacles} are all represented. It is intuitive that these features are important for having basic skill in Mario. The remaining features are important, but only for fine tuning policies. 

\section{Discussion}
The strengths of our approach is its simplicity and generality combined with fast convergence. Once we sample the domain, PCA can be ran on these samples. We can use the transforms given by PCA in the learning cycle whenever we compute a new state. This is a simple matrix operation and adds no additional computational cost. This preliminary work shows that learning can be efficiently performed in these low dimensional spaces at an increased convergence rate. 

\begin{figure}[h!]
  \centering
      \includegraphics[width=0.75\textwidth]{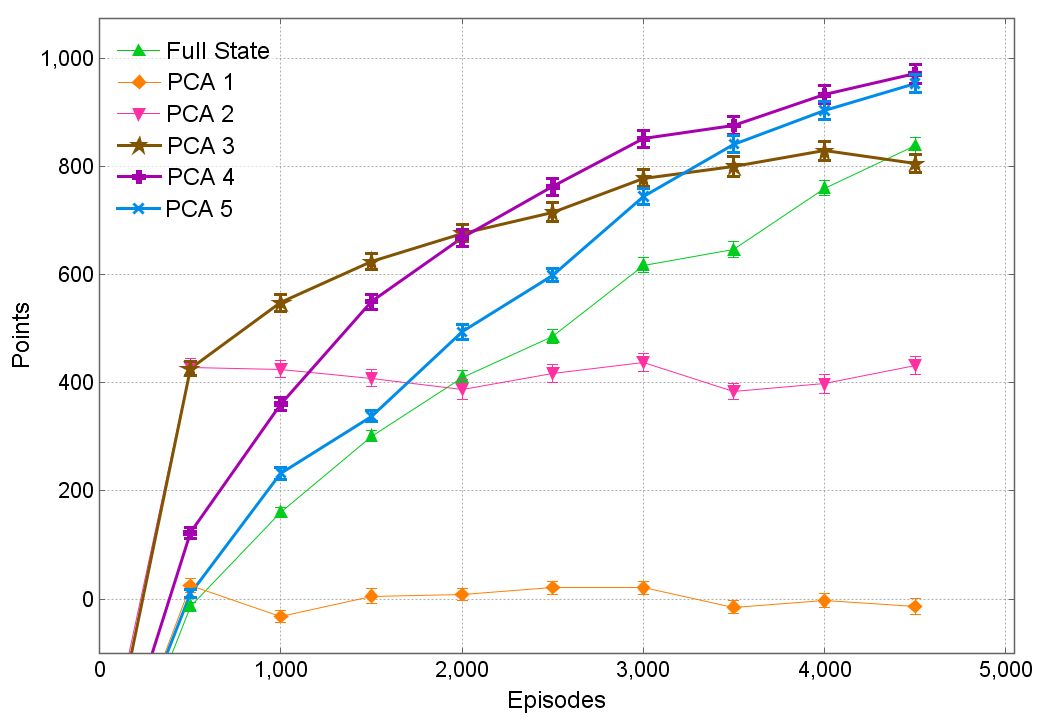}
  \caption{Results with varying manifolds. Lines in bold are experiments that performed better or equal to learning with the full state. Dimensions greater than 5 performed similarly to the full state space, and were not included for clarity. Error bars are over 100 statistical runs.}
  \label{fig:pca_results}
\end{figure}

But, there are some fundamental issues. Our algorithm assumes that all states are important. It is possible for a state to be unnecessary to learning, but have high variance. We also stop learning once we converge in the lower-dimensional space. This ensures that given infinite time, the higher dimensional space will converge to better performance. We discuss our solution to these problems in Section \ref{Future Work} by using an iterative learning approach. 

\section{Future Work}
\label{Future Work}
There are many areas of future work this approach can take. The next step we will take would be to perform additional analysis on the current results. Further experiments include varying the amount of training data given to PCA, or to use random demonstration data. We can then see how the quality and quantity of data affects learning.

A major focus of future work is to improve upon the converged performance of the low-dimensional state representations. Work by \citet{Grzes::mixed} has shown that mixed resolution function approximation works well in complex domains.  They initially learned with a less expressive function approximation to provide early guidance during learning. They then learned using a more expressive function approximation to learn a high quality policy. We can leverage a similar idea. We propose in future work that we can learn until convergence in a $n$ dimensional space. As shown in this work, the convergence time is low.  We then transfer that knowledge to an $n+1$ dimensional space. We hypothesize that this learning technique will converge faster than learning entirely in the full dimensional space.
 

\bibliographystyle{plainnat}
\bibliography{../thesis}

\end{document}